\documentclass[review]{elsarticle}

\usepackage{hyperref}
\usepackage{listings}
\usepackage{xcolor}
\usepackage{caption}
\usepackage{geometry}
\usepackage{float}
\usepackage{amsmath}
\usepackage{amsmath,amssymb,amsthm}
\newtheorem{example}{Example}[section]
\newtheorem{theorem}{Theorem}[section]

\usepackage[export]{adjustbox}
\usepackage{array}
\newcolumntype{P}[1]{>{\centering\arraybackslash}p{#1}}
\newcolumntype{M}[1]{>{\centering\arraybackslash}m{#1}}
\usepackage{tikz}
\usetikzlibrary{topaths}
\usetikzlibrary{arrows}
\journal{Compelexity}

\biboptions{authoryear}

\begin{document}
	
	\begin{frontmatter}
		
		\title{Novel ANN method for solving ordinary and fractional\\ Black-Scholes equation}

		\author[1,2]{Saeed Bajalan}
		%%\ead{saeedbajalan@ut.ac.ir}
		\author[3]{Nastaran Bajalan}
		%%\ead{n.bajalan@tue.nl}
		%%\cortext[mycorrespondingauthor]{Please address all correspondence to saeedbajalan@ut.ac.ir}
		
		\address[1]{PhD in finance, University of Tehran, Tehran, Iran}
		\address[2]{Senior Researcher, Utrecht University, Utrecht, Netherlands}
		\address[3]{PDEng in Software Technology, Eindhoven University of Technology, Eindhoven, Netherlands}
		\begin{abstract}
			The main aim of this study is to introduce a 2-layered Artificial Neural Network (ANN) for solving the Black-Scholes partial differential equation (PDE) of either fractional or ordinary orders. Firstly, a discretization method is employed to change the model into a sequence of Ordinary Differential Equations (ODE). Then each of these ODEs is solved with the aid of an ANN. Adam optimization is employed as the learning paradigm since it can add the foreknowledge of slowing down the process of optimization when getting close to the actual optimum solution.  The model also takes advantage of fine tuning for speeding up the process and domain mapping to confront infinite domain issue. Finally, the accuracy, speed, and convergence of the method for solving several types of  Black-Scholes model are reported.  
		\end{abstract}
		
		
	\end{frontmatter}

	\section{Introduction}\label{sec1}
	Both PDEs of ordinary and fractional order play an important role in pricing of financial derivatives.
	PDEs of ordinary order are the basis of various models proposed for pricing of different types of options. On the other hand, financial markets show fractal behavior (Mandelbrot, 1963; Peters, 1989; Li and Ma, 2005; Huang, Li and Xiong, 2012) and Fractional PDEs (FPDE) which can better reflect the reality of them have gained a lot of interest recently. Hence finding an accurate and efficient approach for solving both types is a critical issue in pricing. 
	
	The most famous PDE in finance is the Black-Scholes (B-S) model, which is broadly adopted for option pricing. So far, studies have presented different approaches for finding numerical solution of this model and its variation when the exact form does not exist (\cite{rad2015,Farnoosh,Farnoosh2016,Farnoosh2017,Golbabai2017a,Golbabai2017b,Rashidinia2017a,Rashidinia2017b,Sobhani}).
	
	Furthermore, in order to depict the fractal structure in financial market, the classical B-S equation has been generalized by means of fractional derivatives which has the property of self-similarity and better modeling of long-range dependency (\cite{Wyss,Bjork,Meerschaert}). Although the fractional B-S is a powerful tool for explaining the hereditary and memory characteristics of financial market, It is considerably difficult to obtain an accurate solution for it, due to the memory trait of fractional derivatives (\cite{Golbabai2019}).  As a result, numerous researchers have tried techniques for approximating such problems. Among different analytical models presented so far to solve time fractional B-S, the most cited articles employed integral transform (\cite{ChenXu2015a,Kumar2012,Wyss,Jumarie,Li2009,Liang}), wavelet based hybrid methods (\cite{Hariharan}), the separation of variables (\cite{Chen}), the homotopy analysis and homotopy perturbation methods (\cite{Elbeleze,Kumar2014,Kumar2016}), Fourier Laplace transform (\cite{Duan}). However, due to the high computational complexity of these solutions, numerical methods are often better alternatives for solving such mathematical models. \\
	Cartea and del Castillo-Negrete used Backward difference technique together with shifted Grunwald–Letnikov scheme to solve the spatial fractional FMLS process \cite{Cartea}.
	Investigation of Convergence Analysis and comparison of the solution of three space fractional B-S mode were provided by Marom and Momoniat (2009)\cite{Marom}.  Finite differences methods were used to provide a solution for time fractional B-S model in 2013 by Song and Wang (\cite{song2013}). Zahng et al.  also proposed a second order finite difference method in the following year (\cite{Zhang2014}).  Moreover in 2017 weighted finite difference method was utilized to find the numerical solution of the model (\cite{Koleva}). Chen et. al.  investigated Predictor–corrector approaches for the solution of American option in 2015 (\cite{ChenXu2015b}). Khan and Ansari (\cite{ansari}) were the first to use the Sumudu transform to solve the fractional model of European options. In the same year, Zhang provided an unconditionally stable implicit numerical scheme for the model by changing the Riemann-Liouville derivatives to Caputo derivative (Zhang et al., 2016b). As RBF methods are widely used for approximating the ordinary PDEs, the meshless method was proposed for finding the solution of time fractional method as well (\cite{Golbabai2019}). In 2020, the fractional model was numerically approximated by using Quintic and Sixtth order B-Spline functions as the basis for a collocation method providing high accuracy for the generalized B-S mode (\cite{Pradip,ROUL2020472}). In order to provide to sixth order accuracy in solving generalized Black–Scholes model,  Roul and Goura used both of Crank–Nicolson scheme and sextic B-spline collocation method.
	
	Regarding machine learning and ANN in particular, ANN has been traditionally used for predicting option prices (\cite{Malliaris,Yao,Amornwattana,Andreou,Jang,Ozbayoglu}), however, to the best of our knowledge except few researches there is no available literature for solving B-S differential equation by ANN (\cite{Cervera,Eskiizmirliler}). Even these few researches are limited to integer order PDEs and there is no study for solution of fractional order B-S. 
	In this paper we are going to construct a 2-layered Artificial Neural Network (ANN) to solve the Black-Scholes model of either fractional or ordinary orders. The Adaptive Moment Estimation (Adam), which has been specifically created to be used by neural networks, acts as the optimizer in the ANN to add the foreknowledge of slowing down when getting close to the optimal solution. To make the process faster and increase the efficiency of the method Fine tuning is applied to the model.  Also to overcome the problem of infinite problem domain for approximation domain mapping is used to shift the whole problem to a finite interval.
	The rest of paper is organized as follows: section. \ref{sec:black} is dedicated to problem formulation section. \ref{sec:methology} explains the methodology and section. \ref{sec4} presents the numerical results and finally the conclusion is provided in section. \ref{sec5}.

	\section{Problem Formulation} \label{sec:black}
	
	A put option on an underlying asset $S$, is said to follow a Geometric Brownian Motion (GBM), where $\sigma$, $r$ and $W$ are the volatility, interest rate, and Brownian motion respectively, if it obeys the stochastic differential equation as follows:
	\begin{equation}
	\label{eq:1}
	dS=rSdt +\sigma S dw
	\end{equation}
	Using Eq. \eqref{eq:1} and risk-neutral valuation formula together with the classic Feynman-Kac formula the Black-Scholes operator is formed as below:
	\begin{equation}
	\label{eq:blackOrdinary}
	LU(S,t)=-\frac{\partial U(S,t)}{\partial t}-\frac{\sigma^2}{2} S^2 \frac{\partial^2 U(S,t)}{\partial S^2}-rS\frac{\partial U(S,t)}{\partial S}+rU(S,t)
	\end{equation}
	$U(S,t)$ is the unknown function which determines the option price \cite{monro}. It has been specified that this option will have a certain payoff at a certain date in the future, depending on the value(s) taken by the stock up to that date.\\
	
	It is well-known now, that a time-fractional Black-Scholes equation with the derivative of real order $\alpha$ can be obtained to describe the price of an option under several circumstances such as when the change in the underlying asset is assumed to follow a fractal transmission system. Fractional derivatives, as they are called, were introduced in option pricing in a bid to take advantage of their memory properties to capture both major jumps over small periods of time and long-range dependencies in markets. Therefore, the fractional Black-Scholes model can be formulated as follows:
	
	\begin{equation}
	\label{eq:blackFractional}
	LU(S,t)=-{D^\alpha U(S,t)}+\gamma_1\frac{\partial^2 U(S,t)}{\partial S^2}+\gamma_2\frac{\partial U(S,t)}{\partial S}+\gamma_3U(S,t)+f(S,t)
	\end{equation}
	$D$ denotes the fractional derivative. $\alpha$ is a real number. $\gamma_i, ~ i=1,2,3$ and $f(S,t)$ are functions dependent on the values of $\sigma$, $r$, and $\alpha$ written using these notations for simplicity. 
	
	\linespread{1.}
	\begin{table}[h]
		\centering
		\small
		
		\setlength\extrarowheight{2.5pt}
		
		\caption{Option parameters and their definitions}
		
		\begin{tabular}{|M{3.1cm}||M{5cm}|}
			
			\hline
			
			$Parameter$ & $Definition$\\ 
			
			\hline
			
			$T$ & Maturity(years) \\ 
			
			$K$  & Strike price \\ 
			
			$S$ & Underlying asset (Stock price)\\
			
			$r$ & Interest rate\\
			
			$D$ & Fractional Derivative \\ 
			
			$\alpha$ & Real valued Derivative Order  \\
			
			$\sigma$ & Volatility \\
			
			$g(S)$& payoff function\\	
			\hline
			
		\end{tabular}
		
	\end{table}
	
	\linespread{1.5}
	Based on the type of the option, the corresponding condition set is as follows:
	
	\begin{equation}
	\label{eq:ini}
	\begin{cases} 
	U(S,T)=g(S) , \\
	U(a,t)=M_a(t),\\
	U(b,t)=M_b(t).\\
	
	\end{cases}
	\end{equation}
	In some cases, e.g. European, $M_b(t)$ moves toward infinity, thus the problem domain is semi-infinite. Some possible strategies are defined in Section. \ref{sec:unb} to overcome this obstacle when pricing.  
	
	\section{Methodology}\label{sec:methology}
	In this section, four main concepts employed in the present approach are explained. Then, the necessary steps to be taken for finding the solution are combined forming the proposed method.

	\subsection{Time Discretization}
	Solving multi-variable equations increases the time complexity and the risk of producing inconsistent answers by computational softwares. In summary, time discretization methods are useful tools for converting such models into a series of Ordinary Differential Equations (ODE).
	This approach is the result of applying finite difference methods on one dimension of an equation to approximately calculate the value of derivatives with respect to that dimension.\\
	Since here fractional equations are also investigated, ordinary and fractional time discretization approaches are discussed.
	Suppose that the Partial Differential Equation (PDE) to be solved is defined on $(S,t)\in [a,b]*[0,T]$. Based on this method $U(S,t)$ in each time-step is defined as $U_i(S)=U(S,i\Delta t)$, where $U(S,t)$ is the answer, $\Delta=\frac{T-0}{N}$, and $N$ is the number of time-steps, then: 
	
	\subsubsection{Ordinary time discretization}
	Consider the PDE to be solved is as follows:
	\begin{equation}
	\frac{\partial U(S,t)}{\partial t}=\varOmega(\frac{\partial U(S,t)}{\partial S},\frac{\partial^2 U(S,t)}{\partial S^2},U(S,t),S,t)
	\end{equation}
	where $U(S,t)$ is the answer and $\varOmega$ is a linear or non-linear function.
	Instead of solving this problem on the two dimensions, it can be converted to a series of dependent ODEs.\\
	By using the defined $U_i(S)$ the following equation is constructed:
	\begin{equation}
	\frac{\partial U_i(S)}{\partial t}=\theta (\varOmega(\frac{\partial U_i(S)}{\partial S},\frac{\partial^2 U_i(S)}{\partial S^2},U_i(S),S,i\Delta t))+(1-\theta) (\varOmega(\frac{\partial U_{i-1}(S)}{\partial S},\frac{\partial^2 U_{i-1}(S)}{\partial S^2},U_{i-1}(S),S,(i-1)\Delta t))
	\end{equation}
	The method is \textit{implicit}, if $\theta =1$. In this case only posterior time-step is used. The method is called \textit{explicit} if  $\theta =0$ where only computing time-step is utilized to approximate the solution. If the value of $\theta$ is equal to $\frac{1}{2}$ the method is the common \textit{Crank–Nicolson} method which is unconditionally stable and of the second order in time and it uses both posterior and computing time-steps for approximating the solution of model. Due to the non-smoothness of the payoff function and the activation functions in our ANN, the Crank–Nicolson can not reach its second order convergence. It can also cause extra inconsistencies because of the same problem. So from this point $\theta=0$ is considered.
	\subsubsection{Fractional Time Discretization}
	
	Suppose the time fractional PDE (FPDE) to be solved is as follows:
	\begin{equation}
	D^\alpha_t U(S,t)=\varOmega(\frac{\partial U(S,t)}{\partial S},\frac{\partial^2 U(S,t)}{\partial S^2},U(S,t),S,t)
	\end{equation}
	where $\alpha$ denotes the Caputo derivatives of the function and $0<\alpha<1$. As the first step for such FPDEs Caputo derivative should be discretized. (Reader are advised to see \cite{mma} for the preliminaries and through information about this derivatives).  Consider the following theorem.
	\begin{theorem}
		Suppose that $[0, T]$ is divided to $N$ parts with step-size of $\Delta t=\frac{T}{N}$, $0<\alpha<1$ and $q(t)\in C^{2}[0,t_k]$ where $t_k = k \Delta t$, the following holds for this interval:
		\begin{equation}
		\begin{split}
		\bigg|\frac{1}{\Gamma(1-\alpha)} \int_{0}^{t_k}\frac{q'(t)}{(t-t_k)^{\alpha}} \Delta t -\frac{\Delta t^{-\alpha}}{\Gamma(2-\alpha)}\bigg[b_0q(x,t_k) - \sum_{m=1}^{k-1}(b_{k-m-1}-b_{k-m})q(t_m) - b_{k-1}q(t_0)\bigg] \bigg|\\\leq \frac{1}{\Gamma(2-\alpha)} \bigg[\frac{1-\alpha}{12} + \frac{2^{2-\alpha}}{2 - \alpha} - (1+2^{-\alpha})\bigg]\max_{0\leq t \leq t_k}|q''(t)|\Delta t^{2-\alpha},
		\end{split}
		\end{equation}
		where $b_m = (m+1)^{1-\alpha} - m^{1-\alpha}$.
	\end{theorem}
	\textit{proof.} See the proof in\cite{suna}.$\qed$
	
	According to the above theorem, the Eq. \eqref{eq:blackFractional} can be discretized in the following form: 
	\begin{equation}
	\mathcal{D}^{\alpha}_{t}U_{n+1}(S) \approx \frac{1}{\Gamma(2-\alpha)} \sum_{m=0}^{n} \frac{\hat{U}_{n+1-j}(S) - \hat{U}_{n-j}(S)}{\Delta t^\alpha} b_m = H[\hat{U}_{n+1}(S)].
	\end{equation}
	
	Now, the unknown function should be approximated in each time-step such that it satisfies Eq. \eqref{eq:blackFractional} and also, its initial and boundary conditions in Eq. \eqref{eq:ini} . In this regard, the boundary conditions can be satisfied by considering $U_N(S) = g(S)$ in computations. On the other hand, to satisfy the boundary condition, the sum of least square error methods is used in Section. \ref{sec4}.

	The remaining step is satisfying the Eq. \eqref{eq:blackFractional}, so for this purpose the cost function is chosen as below:
	\begin{equation}
	Cost(S, W) = \frac{1}{2Nr}\sum_{i=1}^{r}\sum_{n=0}^{N-1}\bigg[\frac{1}{\Gamma(2-\alpha)} \sum_{m=0}^{n} \frac{\hat{U}(S_i, t_{n+1-m}) - \hat{U}(S_m, t_{n-m})}{dt^\alpha} b_m - H[\hat{U}(S_i,t_{n+1})]\bigg]^{2}.
	\end{equation}
	where $S=(S_1, S_2,\dots, S_r)$ and $S_i$ is the $i$-th training data. To find the optimum weights for the network this cost function should be minimized subject to the $W$, so the following nonlinear least square problem is obtained:
	\begin{equation}
	\min_{W} \frac{1}{2Nr}\sum_{i=1}^{r}\sum_{n=0}^{N-1}\bigg[\frac{1}{\Gamma(2-\alpha)} \sum_{m=0}^{n} \frac{\hat{U}_{n+1-m}(S_i) - \hat{U}_{n-m}(S_m)}{dt^\alpha} b_m - H[\hat{U}_{n+1}(S_i)]\bigg]^{2}.
	\end{equation}
	It is noteworthy that when $\alpha=1$, the ordinary time-discretization method will be used.
	
	\subsection{Function Approximation}
	According to the universal approximation theorem, every continuous function can be approximated by a feed-forward neural network \cite{universal}. This theorem states that any linear function can be approximated by an ANN without any hidden layers. But for functions of higher orders, the approximation can be well-established, if the ANN has at least one hidden layer.
	To calculate the price of an option based on Eq. \eqref{eq:blackOrdinary} and Eq. \eqref{eq:blackFractional}, the value of option at each time-step using a 2-layered network should be approximated as follows:
	\begin{equation}
	N(W,S)=\varPsi(V \varphi(W  S+B_0)+B_1)
	\end{equation}
	Where $n$ is the number of neurons in the hidden layer, $B_0=\{ b_1,b_2,\dots, b_n\}$, $B_1=\{\beta_1\}$, $\varphi_i,~ i\in{1,2,\dots,n}$ are the activation functions in the hidden layer, and $\varPsi$ is the activation function for the output layer. 
	The above formula can be seen in Figure. \ref{fig:network}.\\
		\begin{center}
		
		\begin{tikzpicture}
			[   cnode/.style={draw=black,fill=#1,minimum width=3mm,circle},
			]
			\node[cnode=red,label=0:$U$] (s) at (6,-2.5) {$\varPsi$};
			\node[cnode=blue,label=180:$s$] (x-1) at (0,-2.5) {};
			\node[cnode=yellow,label=90:$1$] (x-0) at (0,0) {};
			\node[cnode=yellow,label=90:$1$] (p-0) at (3,0) {};
			
			\node at (3,-4) {$\vdots$};
			
			\draw (p-0) -- node[above,sloped,pos=0.3] {$\beta_1$} (s);
			\foreach \x in {1,...,4}{
				\pgfmathparse{\x<4 ? \x : "n"}
				\node[cnode=gray] (p-\x) at (3,{-\x-div(\x,4)}) {$\varphi_{	\pgfmathparse{\x<4 ? \x : "n"} \pgfmathresult}$};
				\draw (p-\x) -- node[above,sloped,pos=0.3] {$v_{\pgfmathresult}$} (s);
			}

			\foreach \y in {1,...,4}
			{
				\pgfmathparse{\y<4 ? \y : "n"}
				\draw (x-0) -- node[above,pos=0.3] {${ b_\pgfmathresult}$} (p-\y);
			}
			\foreach \y in {1,...,4}
			{
				\pgfmathparse{\y<4 ? \y : "n"}
				\draw (x-1) -- node[above,sloped,pos=0.5] {${ w_\pgfmathresult}$} (p-\y);
			}
			
		\end{tikzpicture}
		\begin{figure}[h]
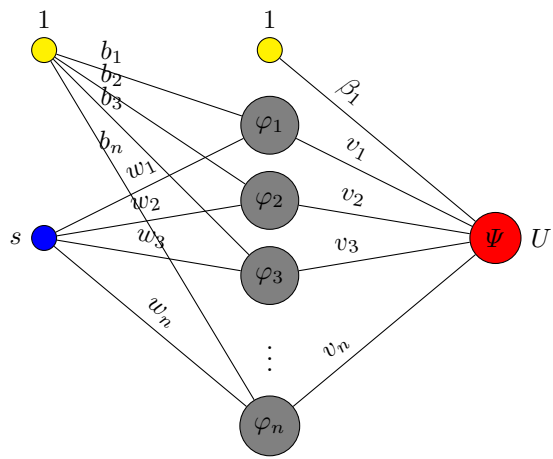

			\caption{The topology of the network used for solving Black-Scholes model.}
			\label{fig:network}
			
		\end{figure}
	\end{center}
	The nodes commensurate with the edges $b_i$ and $\beta_1$ in this network are called the biases whose inputs are unchangeable $1$s and their weights are additional parameters as means of adjusting the output of the next layer. In other words, they help the network fit best for the given data. 
	
	The most famous activation function in deep learning and neural networks is sigmoid. 
	The two main reasons the sigmoid function is widely used are the derivatives of it and its value. The sigmoid function defined as follows and its values are in $[0,1]$ :
	\begin{equation}
	sigmoid(x)=\frac{1}{1+\exp(-x)}
	\end{equation}
	These features make it a perfect candidate for problems that produce probabilities and for the Black-Scholes model. The values of options are nonnegative and the effect of other parameters will not increase the calculations as they are multiplied by smaller values produced by sigmoid. On the other hand, the derivative of this well-known activator, its slope,  is easily calculable between any two arbitrary points:
	
	\begin{equation}
	sigmoid'(x)=\frac{exp(-x)}{(1+\exp(-x))^2}
	\end{equation} 
	%The process of solving the Black-Scholes model using this network is forming an objective function and optimizing its value through learning in the network shown in Figure. \ref{fig:network}. First, the training dataset which is further discussed in \ref{sec:points} are presented to a network initialized by random values. Then during several epochs, the weights are updated until $N(x,x)$ can approximate the value of the Black-Scholes model perfectly.
	Although linear functions such as identity are not favorable for hidden layers, as they take away the chance to generalize and adapt from the network, it is possible to use them as the activator of the output layer hence the hidden layers are present and are directly interacting with inputs.
	
	\subsection{Fine-Tuning}
	One of the key factors in the present study is the possibility of applying the fine-tuning methods during the training process. 
	Fine-tuning is employing a previously trained neural network to find the solution of a new similar task. This process is normally applied to datasets related to images and voices. However, following the same approach, it is possible to increase the accuracy and speed of the network in this work.\\
	Building and validating an ANN from scratch can be a huge task in its own right, depending on what data being trained on it, many parameters such as the number of layers and the number of nodes in hidden layers, the proper activation functions, and learning rate should be found through trial and error.  If a trained model that already does one task well exist, and that task is similar to ours in at least some remote way, then everything the model has already learned can be taken advantage of and applied to the new specific task.
	If the task is completely similar, like what we are facing when solving the problem at different time-steps, the exact weights can be used as the initial values. If the models are somewhat similar, still some knowledge exists on the previous network which is notable for speeding up the process of building/modifying and training the network for the new task. Then the only job remains for the network is learning the new features and properties that were not available in the former task. 
	Here, Once the network is trained for the first time-steps, the obtained parameters, weights, and biases can be effectively reemployed for training the data fed to the network in other time-steps.
	The approximated solution and its convergence rate are compared in section. \ref{sec5}.
	
	\subsection{Domain Mapping} 
	\label{sec:unb}
	Considering the vulnerability of neural networks due to the bounded domain of their activation functions in calculations on infinite domains, the domain mapping approach is utilized to shift the problem from its semi-infinite domain to a finite interval. This helps to prevent the error caused by common solutions such as truncation of the domain.
	
	%Although many options are defined on finite domains, some other options such as American and European ones are defined on semi-infinite intervals. 
	On finite domains, $S=x$ will be considered, but in semi-infinite domains, transformation formulas should be used for shifting the problem to a desired finite one. Here $x(S)=\frac{2}{\pi}\arctan(\frac{S}{L})$ is used to shift the problem's domain which is $[0,\infty)$ to $[0,1]$, in which $L$ is the characteristic length of the mapping \cite{boyd}.
	
	Here, $L$ is chosen in a way that $60\%$ of all training points stand before the mapped strike price because the price significantly differs from zero in $[0, K]$ (\cite{amaniPricing}). It means that by defining $l$ as an indicator for $0.6$, in the view of the fact that these points are equidistant, $L$ is computed as follows:
	$$L=\frac{K}{\tan(\frac{\pi}{2}l)}$$
	
	First, let us introduce the following notations:
	
	\begin{equation*}
	U(S,t)=\tilde{U}(x,t),~~~	S=L \tan(\frac{\pi}{2}x)
	\end{equation*}	
	\begin{equation*}
	\varUpsilon \triangleq \frac{\partial S}{\partial x}=\frac{L\pi}{2\cos^2(\frac{\pi}{2}x)},\quad \varTheta \triangleq \frac{\partial \varUpsilon^{-1}}{\partial x}=-\frac{2\cos(\frac{\pi}{2}x) \sin(\frac{\pi}{2}x)}{L}
	\end{equation*}
	
	Hence the derivatives needed for the calculations according to Section. \ref{sec:black} are:
	
	\begin{equation*}
	\label{eq:fide}
	\frac{\partial {U}(S,t)}{\partial S}=\frac{\partial \tilde{U}(x,t)}{\partial x} \frac{\partial x}{\partial S}= \frac{1}{\varUpsilon}\frac{\partial \tilde{U}(x,t)}{\partial x}
	\end{equation*}
	
	\begin{equation}
	\label{eq:sedeo}
	\frac{\partial^2 {U}(S,t)}{\partial S^2}=\frac{\partial}{\partial S}(\frac{\partial {U}(S,t)}{\partial S})
	=\frac{1}{\varUpsilon^2} \frac{\partial^2 \hat{U}(x,t)}{\partial x^2} + \frac{\varTheta}{\varUpsilon} \frac{\partial \hat{U}(x,t)}{\partial x} 
	\end{equation}
	
	By Substituting Eq. \eqref{eq:sedeo} in the Eq. \eqref{eq:blackFractional} and Eq. \eqref{eq:blackOrdinary} the domain of the obtained Black-Scholes model is $[0,1]$.
	
	Since the transformation is applied to the whole problem, the payoff function, and boundary conditions of Eq. \eqref{eq:ini} should be changed as well.
	\subsubsection{Discussion}
	Using Domain Mapping helps to approximate the answer on the whole interval of the problems. On the other hand, the number of training data points needed for solving the model on smaller intervals is significantly less as can be seen in \ref{sec5}. However, it can decrease the accuracy since the whole domain is being compacted into one small domain and loss of information may occur. Meaning that truncating the domain causes a perfect approximation on a sub-domain of problems but with acceptance of a bit of loss in accuracy the whole domain can be covered. So there lays a trade-off between these two methods. 
	
	\subsection{Adaptive Moment Estimation Learning}
	
	\label{sec:4}
	
	Regression modeling is used to determine coefficients of mathematical functions based on empirical data. The method of least squares determines the coefficients such that the sum of the squares of the deviations between the data and the curve-fit is minimized. Finding a satisfactory solution to nonlinear least-square problems is one of the famous topics among scientists who work on nonlinear systems of equations.
	For minimizing a vector function, $\|\varLambda(x)\|$, that is defined as $\varLambda:\mathbb{R}^n\rightarrow \mathbb{R}^m$, and $m\geq n$ with respect to a predefined $x=(x_1,x_2,...,x_n)$, that is to say, $x^*\in\mathbb{R}^n$ is found in a way that
	\begin{equation}\label{avali}
	\varUpsilon(x)=\frac{1}{2}\sum_{i=1}^m (\varLambda_i(x))^2=\frac{1}{2}\|\varLambda(x)\|^2=\frac{1}{2}f^T(x).f(x)
	\end{equation}
	in which
	\begin{equation}\nonumber
	x^*=\min_x \{\varUpsilon(x)\}.
	\end{equation}
	Several methods have been introduced for solving this nonlinear least square model so far. As we can see the same vector can be constructed when solving differential equations. Similarly, For the Black-Scholes models, in each time-step the final goal is finding the proper network weights solving the following system of equations:
	
	\begin{equation}
	eq_1=U(S,t_i)-\hat{U}(S,t_i)=0, ~~ i=0\dots M,
	\end{equation}
	\begin{equation*}
	eq_2=\hat{U}(0,t_i)-f(S)=0, ~~ i=0\dots M,
	\end{equation*}
	\begin{equation*}
	eq_3=\lim_{x\to boundary}(\hat{U}(S,t_i)-g(S))=0, ~~ i=0\dots M,
	\end{equation*}
	
	The best possible solution to this system of equations is calculated when the sum of the squares of these equations gets smaller. Hence the problem is converted into an optimization problem. By minimizing the following equation, the appropriate weights and biases for our network are found:
	
	\begin{equation}
	objective(\hat{U}(S,t))=\min_{W,V,B_1,B_2}eq_1^2+eq_2^2+eq_3^2
	\end{equation} 
	
	%Different optimizers are available to be used as the learning paradigm in neural networks.
	Gradient Descent ($GD$) is the most famous iterative algorithm employed as a learning paradigm to solve regression problems. In $GD$, After initializing the weights the gradients, $G$, of the cost function is calculated. The cost function is the sum square error of the output based on the desired output for each member of the training dataset. Then based on $G$ the weights become updated 
	$W = W - \eta G$. This process is repeated by considering the new values as the initial ones until the cost function is desirably minimized.
	$\eta$ is known as learning parameter is used to balance the rate of increase or decrease in each iteration. It should be noted that the only constraint on this value is $0<\eta<1$. Various methods and theorems are introduced to find the boundaries of this value according to the problem.  Some suggest using a big value and decreasing its value as the result approaches the correct value. In contrast, some others suggest choosing a very small value and then increasing it exponentially in time when the correct direction is found. But generally, they all prefer making this parameter a function of time.  While none of them proposes a single formula to calculate the exact amount of it for the best approximation \cite{cyclic,darken,monro}.

	GD family has different optimizers such as Stochastic Gradient Descent (SGD), Adaptive Moment Estimation(Adam),  Root Mean Square Propagation (RMSprop), and Nesterov Accelerated Gradient (NAG) which are mostly used in deep learning because of their speed and strength according to various control parameters such as the size of the training datasets and the pattern in which the training data is scattered. %However, they are mostly rooted in gradient descent.  Gradient descent is the most common method used to optimize deep learning networks. First proposed in the 1950s, the technique can update each parameter of a model, observe how a change would affect the objective function, choose a direction that would lower the error rate, and continue iterating until the objective function converges to the minimum.
	%Essentially Adam is an algorithm for gradient-based optimization of stochastic objective functions. It combines the advantages of two SGD extensions — Root Mean Square Propagation (RMSProp) and Adaptive Gradient Algorithm (AdaGrad) — and computes individual adaptive learning rates for different parameters.
	
	In GD for updating only one parameter, all available samples in the dataset should be visited, however in SGD, \cite{sgd}, mini-batches which are small subsets of the whole dataset are used to update a single parameter. For relatively large datasets, this causes the algorithm to converge faster. GD is an actual optimizer trying to find the exact gradients while in SGD as explained the algorithm only approximates the gradients and not the precise value. Since SGD fluctuates a lot, due to frequent updates with high variance, it shows a paradoxical behavior. It can explore new and potential directions to find the minimum but at the same time, this behavior puts the network in danger of completely missing the local or global minimum. 
	
	%To overcome the problem of these heavy fluctuations, learning algorithms with momentums where proposed \cite{}. Adding momentum causes the network to have the foreknowledge of slowing the process of getting closer to the optimum point down when it is close enough to the answer. The algorithm starts by making big changes in all or a mini-batch of the dataset. Then using the expected value, based on momentums makes a correction in the direction or the value of this change. \\
	A solution was proposed by the father of propagation, Geoffrey Hinton \cite{father}. This through study which has not been academically submitted or published gained a lot of attention. The proposed algorithm fights the possibilities of vanishing or exploding the gradients. In other words, RMSprop normalizes the gradient using a moving average of squared gradients. This normalization balances the step size, reducing the step-size for large gradients to avoid exploding and increasing it for small ones to avoid vanishing. Since this approach uses the exponentially decaying average, it is related to the most recent gradients, so the past gradient would not play a great role in updating the parameters. This leads to slow changes in the learning rate; however, it is relatively faster than GD. 
	\\
	%The optimizers are promising for many problems, but still, some problems need the parameters to have their individual changes so they need distinctive learning rates.
	So far, it can be seen that RMSProp and SGD are the best options. Adam is an adaptive algorithm which is generally considered as the combination of these two paradigms with momentum. This methodology has been specifically created to be used by neural networks.\\ 
	Like RMSprop, Adam employs squared gradients to modify the learning rate. Also, the first and second moments are utilized using the moving average of the gradient like SGD. However, a specific learning rate is calculated for each network parameter (weights) using two hyper-parameters.  Here a summary of how Adam optimization works for the present model is provided. For the complete explanation, readers are encouraged to study \cite{Adam}.
	The convergence of the method has been described in several great papers. But finally, all of the studies confirm the convergence proof provided in the first papers. \cite{Reddi2018OnTC}.\\
	All computations are done using\textit{ autograd} package of Python. In this package Adam optimization is implemented as follows:
	\linespread{0.95}
	\begin{lstlisting}
	def adam(grad, w, callback=None, num_iters=100,
	step_size=0.001, b1=0.9, b2=0.999, eps=10**-8):
	m = np.zeros(len(w))
	v = np.zeros(len(w))
	for i in range(num_iters):
	g = grad(w, i)
	if callback: callback(w, i, g)
	m = (1 - b1) * g      + b1 * m  # First  moment estimate.
	v = (1 - b2) * (g**2) + b2 * v  # Second moment estimate.
	mhat = m / (1 - b1**(i + 1))    # Bias correction.
	vhat = v / (1 - b2**(i + 1))
	x = x - step_size*mhat/(np.sqrt(vhat) + eps)
	return x
	\end{lstlisting}
	\linespread{1.5}
	As the name of the method describes, it is derived from adaptive moment estimation. $n$-th moment of a random variable is defined as the expected value of that variable to the power of n:
	\begin{equation}
	m_n=E[w^n]
	\end{equation}
	where $m$ shows the $n$-th moment and $w$ is a random variable.
	The gradient of the cost function of the neural network can be considered a random variable since it usually evaluated on some small random batch of data. The first moment is mean, and the second moment is uncentered variance. To estimates the moments, Adam utilizes exponentially moving averages, computed on the gradient evaluated on a current mini-batch:
	\begin{align}
	m_i=\beta_1m_{i-1}+(1-\beta_1)g_i\\
	v_i=\beta_2v_{i-1}+(1-\beta_2)g_i^2
	\end{align}
	$m$ and $v$ denote the moving averages, and g is the gradient of the current data presented to the network. According to \cite{Adam} which is also mentioned the above snippet from \textit{autograd} package, the values of hyper-parameters $\beta_1$ and $\beta_2$ have two default values of 0.9 and 0.999 respectively. While the authors did not discuss the choosing process of these two variables, all studies reported very promising and in most cases perfect estimations using these two default values (see also \cite{hyper}). The vectors of moving averages are initialized with zeros at the first iteration.\\
	The remaining problem with these moments was being biased towards zero since $m_i$ and $v_i$ are initialized as vectors of 0's. In other words, especially during the initial epochs, and when the decay rates are small (i.e. 
	$\beta_1$ and $\beta_2$ are close to 1) the values of $m_i$ and $v_i$ will not change significantly or even at all. So the authors proposed the following bias corrections in order to surmount this obstacle:
	
	\begin{equation}
	\hat{m_i}=\frac{m}{1-\beta_1^i}
	\end{equation}
	\begin{equation}
	\hat{v_i}=\frac{v}{1-\beta_2^i}
	\end{equation}
	
	Now, for each of the parameters (weights) a specific updating rule can be created:
	\begin{equation}
	w=w-\eta\frac{\hat{m}}{\sqrt{\hat{v}+\epsilon}}
	\end{equation} 
	$\epsilon$ is an control parameter preventing the fractional part from producing a division by zero error.\\
	Different scientific studies have shown that Adam outperforms other methods. According to empirical practices, this method has better performance and accuracy. This is also discussed in Section. \ref{sec5}. One problem that is stated by many studies is the convergence of the method. However, \cite{Adam} provided the analysis for the convex problems, other papers argued the convergence of the method on a few non-convex problems. And with some modification, they finally agreed on its usability. 
	
	In \cite{conv}, the full analysis of the convexity of the Black Scholes model is proposed. Due to differences such as the failure of put-call parity in real markets instead of theory, this paper proves that for all American options they preserve their convexity in bubbled markets as well as non-bubbled ones. They showed European options are convexity preserving only for bounded payoffs. Thus, in this respect, the prices of American options are more robust than their European counterparts. In the same study, it is shown that models for bubbles are convexity preserving for bounded contracts. More precisely, consider $(x, t) \in [0,\infty)\times[0, T]$, and let $u1(x, t)$ and $u2(x, t)$ be the option prices such that their corresponding volatilities are nonnegative $\alpha_1$ and $\alpha_2$ which satisfy $\alpha_1(x, t) \leq \alpha_2(x, t)$: 
	\begin{theorem}
		Assume that g is concave. Then $u(x, t)$ is concave in $x$
		for any $t \in [0, T]$. Moreover, the option price is decreasing in the volatility,
		that is, $u1(x, t) \geq u2(x, t)$ for all $(x, t) \in [0,\infty) \times [0, T]$.
		Similarly, if g is convex and bounded, then u(x, t) is convex in x for any
		$t \in [0, T]$. Moreover, the option price is increasing in the volatility, that is,
		$u1(x, t) \leq u2(x, t)$ for all $(x, t) \in [0,\infty) \times [0, T]$.
	\end{theorem}
	The full proof is available in \cite{conv}. As they mentioned the proof is valid under the assumption of the uniqueness of the result for such an option, which is proved in their thorough study on properties of Black-Scholes models in more realistic markets as well.
	
	\section{Numerical results and discussion}\label{sec4}
	In this section, three test examples with exact solutions are chosen according to the previous works for examining the accuracy and efficiency of the proposed ANN. According to what is mentioned in Section. \ref{sec:4}, this approach is applicable to other kinds of options such as barriers and American options. 
	All computations are performed using $Python.3.7$ software on a $2.7 GHz$ $Intel$ $Core ~i7$ CPU machine with $16 Gbyte$ of memory. Only one hidden layer with $20$ hidden neurons is used in all of the samples. The initial weights are scattered in $[-0.01,0.01]$.  The number of epochs for the first time-stamp is $5000$ and for the rest of the steps thanks to Fine-Tuning, decreasing this number to $ 1200$ provides very promising results. All values in $[2000, 7000]$ for the first time-step iterations and all values in $[600, 7000]$ do not lead to overfitting/underfitting, but the best results in our experiment achieved using the stated values. If the learning rate is low, then training is more reliable, but optimization will take a lot of time because steps towards the minimum of the loss function are small and on the other hand, if the learning rate is high, then training may not converge or may even diverge. Weight changes can be so big that the optimizer overshoots the minimum and makes the loss worse. In this research best values for learning rate are found using genetic algorithm.

	\begin{example}\label{test:eu_call}
		\normalfont
		Let us consider a European call option, which its interest rate, volatility and strike price are $0.05$, $0.2$, $10$ respectively. The governing equation is similar to Eq. \eqref{eq:euroex} and the boundary conditions set is as follows:
		\begin{equation}
		\begin{cases} 
		U(S,T)=\max(S-K,0) , \\
		U(0,t)=0,\\
		\displaystyle \lim_{S\to \infty} U(S,t)=S-K\exp(-rt)\\
		\end{cases}
		\end{equation}
		With the maturity of $1$, in years, the approximate price at $t=0$ is shown in Figures. \ref{fig:call_main}.a and \ref{fig:call_main}.b.
		The exact solution of this call option can be obtained using the following analytical solution denoted by $U_{exact}(S,t)$:
		\begin{equation*}
		d_1=\frac{ln(\frac{S}{K})+(r+(\frac{1}{2})\sigma^2)(T-t)}{\sigma \sqrt{T-t}},
		\end{equation*}
		\begin{equation*}
		d_2=d_1-{\sigma \sqrt{T-t}},
		\end{equation*}
		\begin{equation*}
		N(S)=\frac{1}{\sqrt{2\pi}}\int_{-\infty}^{S}\exp(-\frac{1}{2}y^2)dy,
		\end{equation*}
		\begin{equation}
		U_{exact}(S,t)=SN(d_1)-K \exp(-r(T-t))N(d_2).
		\end{equation}
		When truncating the domain at $15$, the cardinal of training dataset is $150$ containing equidistant points scattered between $[0,15]$ and $\eta=0.03$. The number of time-steps, $N$, is $20$ and the average calculation time per each epoch on the above mentioned configuration is  $0.098s$.   
		Figure. \ref{fig:call_main}.a demonstrates that when the problem domain is truncated, the approximate price is behaving fine until it reaches the truncation point which is $15$ in these examples.
		
	\end{example}
	%%%%% European put
	\begin{figure}[H]
		\addtocounter{figure}{1}
		\minipage{0.42\textwidth}
		\includegraphics[width=\linewidth]{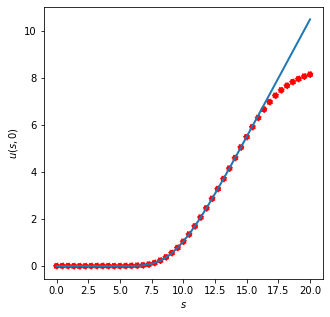}
		\captionsetup{labelformat=empty}
		\caption{(a)}
		\addtocounter{figure}{-2}
		\endminipage\hfill
		\minipage{0.52\textwidth}
		\includegraphics[width=\linewidth]{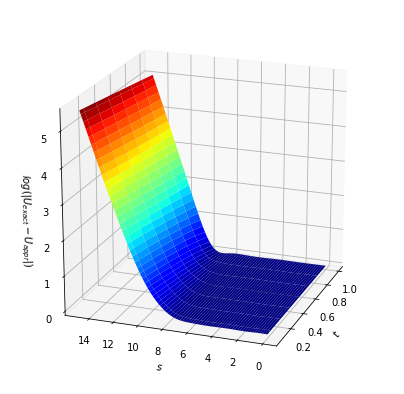}
		\captionsetup{labelformat=empty}
		\caption{(b)}
		\addtocounter{figure}{-1}
		\endminipage\hfill
		\caption{ Plots of the approximated solutions of Example. \ref{test:eu_call} when the number of hidden neurons is $20$ and $dt=\frac{1}{20}$, a)$U(S,0)$  and b) $U(S,t)$, $t \in [0,T]$. }
		\label{fig:call_main}
	\end{figure}
	\begin{figure}[H]
		\addtocounter{figure}{1}
		\minipage{0.450\textwidth}
		\includegraphics[width=\linewidth]{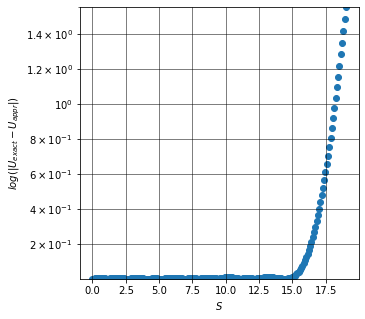}
		\captionsetup{labelformat=empty}
		\caption{(a)}
		\addtocounter{figure}{-2}
		\endminipage\hfill
		\minipage{0.48\textwidth}
		\includegraphics[width=\linewidth]{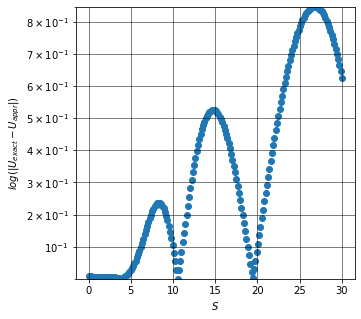}
		\captionsetup{labelformat=empty}
		\caption{(b)}
		\addtocounter{figure}{-1}
		\endminipage\hfill
		\caption{ 	Plots of the logarithmic absolute error for the solution of Example. \ref{test:eu_call} when the number of hidden neurons is $20$ and $dt=\frac{1}{20}$, logarithmic error a)  using truncating approach and b)using mapping function.   }
		\label{fig:call_er}
	\end{figure}
	To solve this issue, the problems is mapped to $[0,1]$ using Section.\ref{sec:unb}, then $U(x,t)$ is computed using the proposed ANN and sigmoid functions as the activation functions, then $U(x,t)$ is reverted to the original model's domain using the inverse mapping function so that $U(S,t)$ is calculated.
	Only $10$ equidistant points are used as training points in this case and the logarithmic absolute errors obtained from two approaches are compared in Figure. \ref{fig:call_er}. It should be noted that after the truncation point the error increases rapidly for the first approach, but when truncating the domain the overall error is higher at the beginning of the interval but it remains steady and even falls at the end of the domain. Since the mapping function converges to infinity on $x=1$. the numerical calculation on software such as Python will not be able to perform the calculations. So these comparisons are done using a very big value for $x=0.9999999$.  The figure confirms the fact that  Adam optimizer performs better as it starts to converge and moves towards the answer in earlier epochs for the first time-step. The average calculation time for SGD and RMSprop are respectively $0.45s$ and $0.63s$ per epoch. 
	It is note worthy that RMSprop crashed due to overflow encounters and the depicted figure is just for comparing, with the learning rate of $0.01$ instead of $0.03$ which somehow might make the comparison unreliable. But the point is observed, this method fails in comparison to the other methods for solving this type of Black-Scholes model. \\
	\begin{figure}[H]
		\addtocounter{figure}{1}
		\minipage{0.34\textwidth}
		\includegraphics[width=\linewidth]{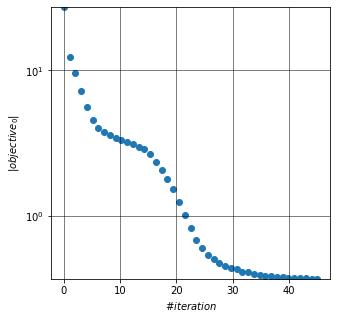}
		\captionsetup{labelformat=empty}
		\caption{(a)}
		\addtocounter{figure}{-2}
		\endminipage\hfill
		\minipage{0.33\textwidth}
		\includegraphics[width=\linewidth]{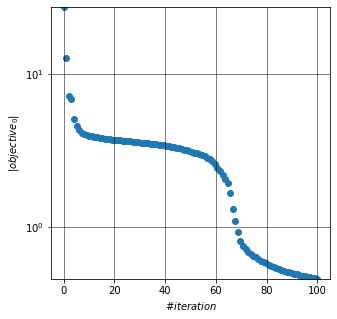}
		\captionsetup{labelformat=empty}
		\caption{(b)}
		\addtocounter{figure}{-1}
		\endminipage\hfill
		\minipage{0.33\textwidth}
		\includegraphics[width=\linewidth]{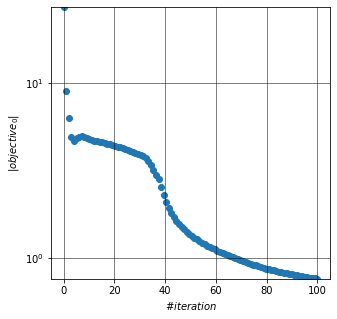}
		\captionsetup{labelformat=empty}
		\caption{(c)}
		\addtocounter{figure}{-1}
		\endminipage\hfill
		\caption{Comparison of networks convergence toward the exact solution of Example. \ref{test:eu_call} using a) Adam optimizer, b) SGD optimizer, and c)RMSprop ($\eta=0.01$)  }
		\label{fig:barr_1_main}
	\end{figure}
	
	Figures. \ref{fig:fine-tuning-call}.a and \ref{fig:fine-tuning-call}.b illustrate the fantastic influence of fine-tuning on the objective convergence. 
	\begin{figure}[H]
		\addtocounter{figure}{1}
		\minipage{0.48\textwidth}
		\includegraphics[width=\linewidth]{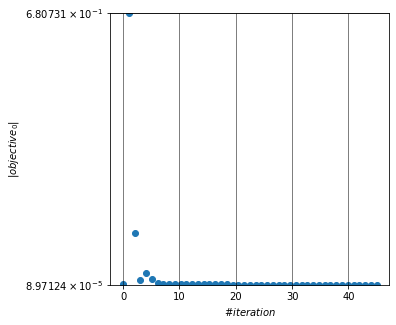}
		\captionsetup{labelformat=empty}
		\caption{(a)}
		\addtocounter{figure}{-2}
		\endminipage\hfill
		\minipage{0.42\textwidth}
		\includegraphics[width=\linewidth]{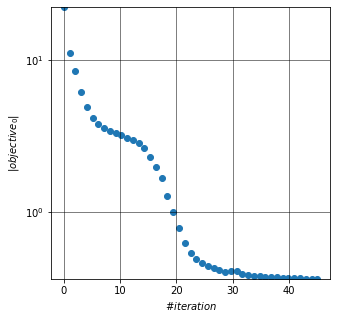}
		\captionsetup{labelformat=empty}
		\caption{(b)}
		\addtocounter{figure}{-1}
		\endminipage\hfill
		\caption{ The effect of fine-tuning on the convergence rate of the method. solving Example. \ref{test:eu_call} with a) Fine-Tuning and  b) Without Fine-Tuning }
		\label{fig:fine-tuning-call}
	\end{figure}

	\begin{example}\label{test:bar1}
		\normalfont
		Consider the following fractional model of a European option with homogeneous boundary conditions as follows:
		\begin{equation}
		D^{\alpha}U(s,t)=a\frac{\partial^ 2 U(s,t)}{\partial t^2}+b\frac{\partial U(s,t)}{\partial t}-cU(s,t)+f(S,t)=0
		\end{equation}

		\begin{equation}
		\begin{cases}
		U(0,t)=t,\\
		U(1,t)=0,\\
		U(S,0)=S^2(S-1).
		
		\end{cases}
		\end{equation}
		The interest rate and volatility are $0.05$ and $0.25$ respectively. Other variables are calculated based on these two variable, $a=\frac{1}{2}sigma^2$, $b=r-a$ and $c=r$. The fractional order of the equation is $\alpha$. Also,
		\begin{equation}
		f(S,t)=(\frac{2t^{2-\alpha}}{\varGamma(3-\alpha)}+\frac{2t^{1-\alpha}}{\varGamma(2-\alpha)})S^2(1-S)-(t+1)^2[a(2-6S)+b(2S-3S^2)-cS^2(1-S)]
		\end{equation}
		The exact solution for this equation is formulated as below when $\alpha$:
		\begin{equation}
		U_{exact}(S,t)=(t+1)^2 S^2(1-S)
		\end{equation}
		The number of time-steps, $N$, is $10$, the number of hidden neurons in this example is $6$, and the average calculation time per each epoch on the above mentioned configuration is  $0.0012s$. Also, $\eta=0.03$.  The approximate solution calculated using the Adam neural network is plotted in Figures. \ref{fig:frac_main}. While smaller number of training data points, $20$ and  $40$, still produced errors in $[0,10^{-2}]$, $60$ equidistant points are used as training points in this case to reach the logarithmic absolute errors illustrated in Figure. \ref{fig:frac_error}.a, giving us the freedom to increase the accuracy even more.
		Because the number of epochs in this example is very small, the absolute errors obtained from SGD, RMSprop, and Adam are compared in Figure. \ref{fig:frac_error}. Here, it can be seen that with a small number of neurons and training points the accuracy of the model is more promising than the other optimizers. The average calculation time for SGD and RMSprop are respectively $0.0059s$ and $0.17s$ per epoch. Figure. \ref{fig:alphas} shows the calculated result at the maturity for different values of $\alpha$. Since the exact solution for other values of $\alpha$ is not available, we can only see that the value of $\alpha$ and the value of the option are directly proportional throughout the whole domain. While the changes are not significant at 2 endpoints of the plot.\\

	\end{example}
	
	%%%%% Barrier first
	\begin{figure}[H]
		\addtocounter{figure}{1}
		\minipage{0.45\textwidth}
		\includegraphics[width=\linewidth]{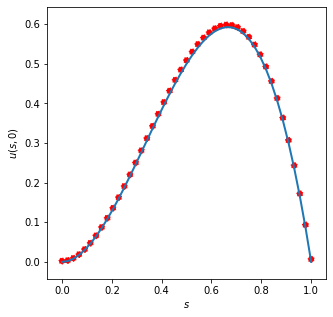}
		\captionsetup{labelformat=empty}
		\caption{(a)}
		\addtocounter{figure}{-2}
		\endminipage\hfill
		\minipage{0.55\textwidth}
		\includegraphics[width=\linewidth]{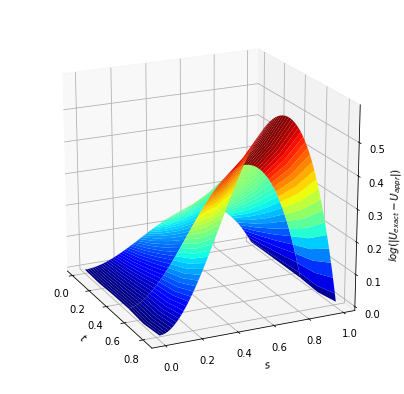}
		\captionsetup{labelformat=empty}
		\caption{(b)}
		\addtocounter{figure}{-1}
		\endminipage\hfill
		\caption{ Plots of the approximated solutions of Example. \ref{test:bar1} when the number of hidden neurons is $60$ and $dt=\frac{1}{20}$, a)$U(S,0)$  and b) $U(S,t)$, $t \in [0,T]$ }
		\label{fig:frac_main}
	\end{figure}
	
	\begin{figure}[H]
		\addtocounter{figure}{1}
		\minipage{0.34\textwidth}
		\includegraphics[width=\linewidth]{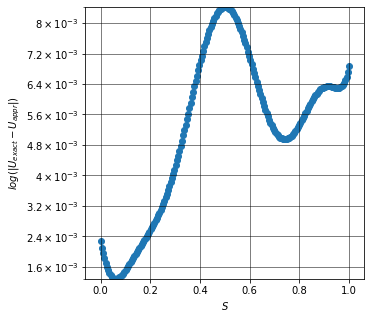}
		\captionsetup{labelformat=empty}
		\caption{(a)}
		\addtocounter{figure}{-2}
		\endminipage\hfill
		\minipage{0.33\textwidth}
		\includegraphics[width=\linewidth]{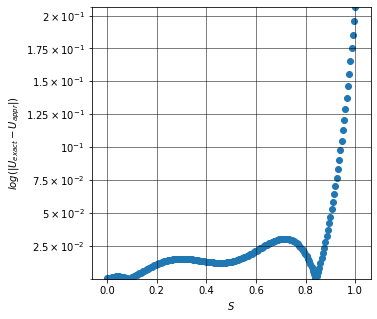}
		\captionsetup{labelformat=empty}
		\caption{(b)}
		\addtocounter{figure}{-1}
		\endminipage\hfill
		\minipage{0.33\textwidth}
		\includegraphics[width=\linewidth]{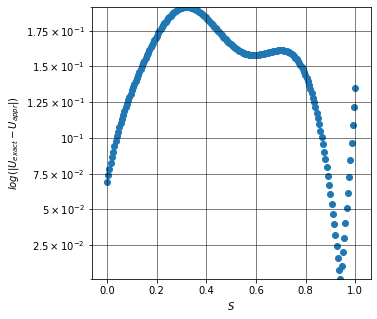}
		\captionsetup{labelformat=empty}
		\caption{(c)}
		\addtocounter{figure}{-1}
		\endminipage\hfill
		\caption{Comparison of networks convergence toward the exact solution of Example. \ref{test:bar1} using a) Adam optimizer, b) SGD optimizer, and c)RMSprop }
		\label{fig:frac_error}
	\end{figure}
	Figures. \ref{fig:fine_tuning_frac}illustrate the fantastic influence of fine-tuning on the objective convergence. When Fine-Tuning is used, the objective function starts with a very small value and hence converge rapidly even for very small values of the cost ($10^{-4}$).
	\begin{figure}[H]
		\addtocounter{figure}{1}
		\minipage{0.48\textwidth}
		\includegraphics[width=\linewidth]{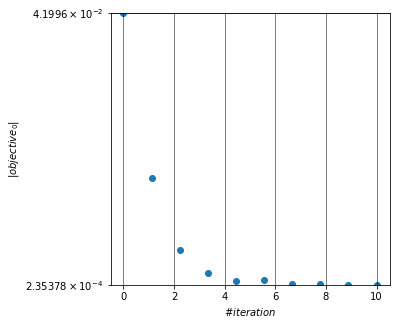}
		\captionsetup{labelformat=empty}
		\caption{(a)}
		\addtocounter{figure}{-2}
		\endminipage\hfill
		\minipage{0.42\textwidth}
		\includegraphics[width=\linewidth]{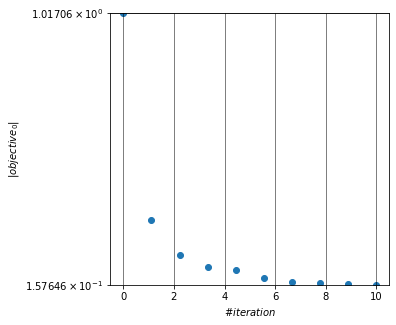}
		\captionsetup{labelformat=empty}
		\caption{(b)}
		\addtocounter{figure}{-1}
		\endminipage\hfill
		\caption{ The effect of fine-tuning on the convergence rate of the method. solving Example. \ref{test:bar1} with a) Fine-Tuning and b) Without Fine-Tuning }
		\label{fig:fine_tuning_frac}
	\end{figure}
	\begin{figure}[H]
		\addtocounter{figure}{1}
		\minipage{0.58\textwidth}
		\includegraphics[width=\linewidth]{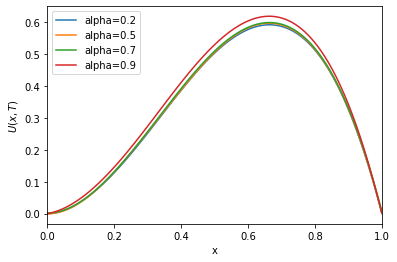}
		\captionsetup{labelformat=empty}
		\caption{(a)}
		\addtocounter{figure}{-2}
		\endminipage\hfill
		\minipage{0.4\textwidth}
		
		\captionsetup{}
		\caption{Plots of the convergence and logarithmic absolute error for the solution of Example. \ref{test:bar1} when the number of hidden neurons is 60 and $dt=\frac{1}{30}$, a) Cost function in each ANN iteration and b) $|U_{exact}(S,T)-U_{app}(S,0)|$. }
		\label{fig:alphas}
		\endminipage\hfill
	\end{figure}

	\begin{example}\label{test:eu_put}
		\normalfont
		Let us consider a European put option, which its interest rate, volatility and strike price are $0.05$, $0.2$, $10$, respectively.
		
		\begin{equation}
		\label{eq:euroex}
		LU(S,t)=\frac{\partial U(S,t)}{\partial t}-\frac{\sigma^2}{2} S^2 \frac{\partial^2 U(S,t)}{\partial S^2}-rS\frac{\partial U(S,t)}{\partial S}+rU(S,t)
		\end{equation}
		\begin{equation}
		\label{eq:eucond}
		\begin{cases} 
		V(S,T)=\max(K-S,0) , \\
		V(0,t)=K\exp(-rt),\\
		\displaystyle \lim_{S\to \infty} V(S,t)=0\\
		\end{cases}
		\end{equation}
		
		With the maturity of $1$, in years, the achieved result is shown in Figure. \ref{fig:call_main_1}.	The exact solution of this put option can be obtained using the following analytical solution denoted by $U_{exact}(S,t)$:
		\begin{equation*}
		d_1=\frac{ln(\frac{S}{K})+(r+(\frac{1}{2})\sigma^2)(T-t)}{\sigma \sqrt{T-t}},
		\end{equation*}
		\begin{equation*}
		d_2=\frac{ln(\frac{S}{K})+(r-(\frac{1}{2})\sigma^2)(T-t)}{\sigma \sqrt{T-t}},
		\end{equation*}
		\begin{equation*}
		N(S)=\frac{1}{\sqrt{2\pi}}\int_{-\infty}^{S}\exp(-\frac{1}{2}y^2)dy,
		\end{equation*}
		\begin{equation}
		U_{exact}(S,t)=-SN(-d_1)+K \exp(-r(T-t))N(-d_2).
		\end{equation}
		
		%%%%% European call
		\begin{figure}[H]
			\addtocounter{figure}{1}
			\minipage{0.4\textwidth}
			\includegraphics[width=\linewidth]{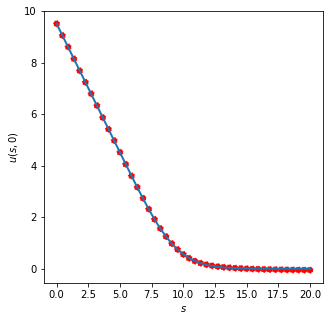}
			\captionsetup{labelformat=empty}
			\caption{(a)}
			\addtocounter{figure}{-2}
			\endminipage\hfill
			\minipage{0.45\textwidth}
			\includegraphics[width=\linewidth]{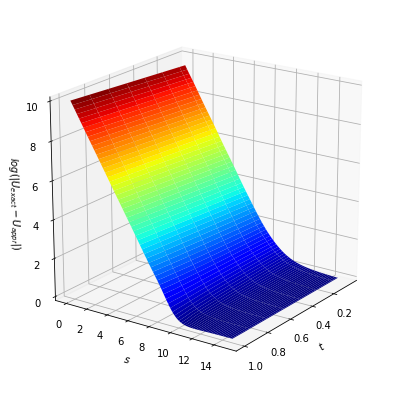}
			\captionsetup{labelformat=empty}
			\caption{(b)}
			\addtocounter{figure}{-1}
			\endminipage\hfill
			\caption{ Plots of the approximated solutions of Example. \ref{test:eu_put} when the number of hidden neurons is $60$ and $dt=\frac{1}{20}$, a)$U(S,0)$  and b) $U(S,t)$, $t \in [0,T]$ }
			\label{fig:call_main_1}
		\end{figure}
		Since the problem domain is unbounded according to the boundary conditions, When truncating the domain at $15$, the cardinal of the training dataset is $110$ containing equidistant point scattered between $[0,15]$ and $\eta=0.2$. The number of time-steps, $N$, is $10$ and the average calculation time per each epoch on the above-mentioned configuration is  $0.032s$. Increasing the number of data points will increase the accuracy for this configuration slightly (other parameters might need to be adjusted as well), however, we preferred this size to reduce complexity and memory usage.

		In Figure. \ref{fig:call_main_1}.a,it is shown that when the problem domain is truncated,unlike Example. \ref{test:eu_put}, the approximate price is behaving fine throughout the whole unbounded interval. But this does not state that this behavior is the expected behavior of the option considering that the boundary condition makes the option price move towards zero. To make it clearer, the errors for truncated and mapped approximate solutions are compared in \ref{fig:call_er_1}. In \ref{fig:call_er_1}.a the absolute error before the truncation point is relatively better than \ref{fig:call_er_1}.b. 
		However, after the truncation point, it starts to increase and then again flattens out which is predictable according to the boundary condition of this specific function. In other words that this behavior can not be generalized to other options as well, because farther points are outside training dataset and network can not learn their values. So the preferred way is employing a mapped domain with lower accuracy but more stable behavior. Besides, only $10$ equidistant points are used as training points in this case. 
		\begin{figure}[H]
			\addtocounter{figure}{1}
			\minipage{0.450\textwidth}
			\includegraphics[width=\linewidth]{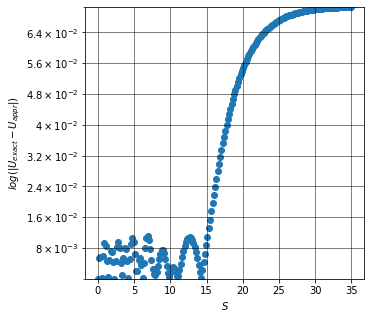}
			\captionsetup{labelformat=empty}
			\caption{(a)}
			\addtocounter{figure}{-2}
			\endminipage\hfill
			\minipage{0.48\textwidth}
			\includegraphics[width=\linewidth]{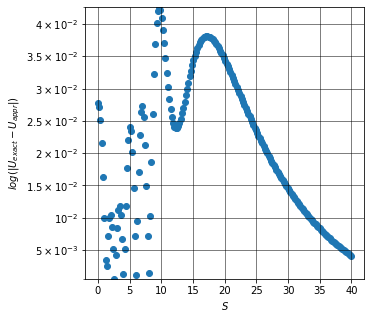}
			\captionsetup{labelformat=empty}
			\caption{(b)}
			\addtocounter{figure}{-1}
			\endminipage\hfill
			\caption{ 	Plots of the convergence and logarithmic absolute error for the solution of Example. \ref{test:eu_put} when the number of hidden neurons is 60 and $dt=\frac{1}{30}$, a) Cost function in each ANN iteration and b) $|U_{exact}(S,T)-U_{app}(S,0)|$.   }
			\label{fig:call_er_1}
		\end{figure}

		\begin{figure}[H]
			\addtocounter{figure}{1}
			\minipage{0.34\textwidth}
			\includegraphics[width=\linewidth]{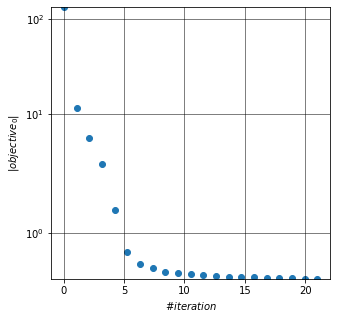}
			\captionsetup{labelformat=empty}
			\caption{(a)}
			\addtocounter{figure}{-2}
			\endminipage\hfill
			\minipage{0.33\textwidth}
			\includegraphics[width=\linewidth]{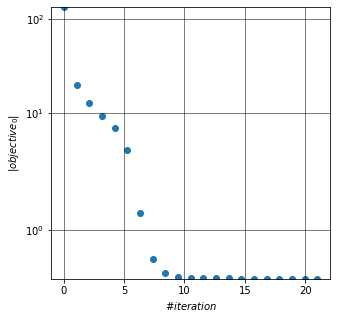}
			\captionsetup{labelformat=empty}
			\caption{(b)}
			\addtocounter{figure}{-1}
			\endminipage\hfill
			\minipage{0.33\textwidth}
			\includegraphics[width=\linewidth]{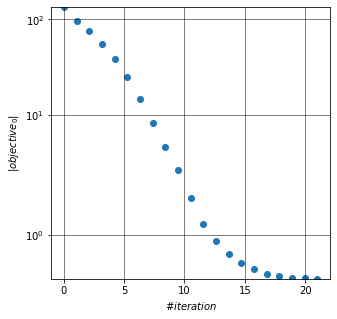}
			\captionsetup{labelformat=empty}
			\caption{(c)}
			\addtocounter{figure}{-1}
			\endminipage\hfill
			\caption{Comparison of networks convergence toward the exact solution of Example. \ref{test:eu_put} using a) Adam optimizer, b) SGD optimizer, and c)RMSprop }
			\label{fig:optimizers}
		\end{figure}
		\begin{figure}[H]
			\addtocounter{figure}{1}
			\minipage{0.48\textwidth}
			\includegraphics[width=\linewidth]{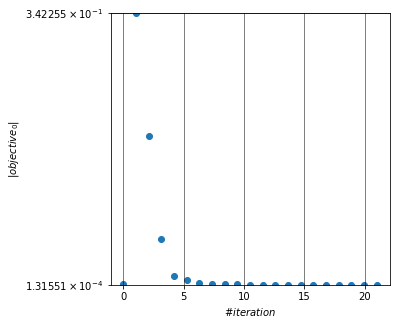}
			\captionsetup{labelformat=empty}
			\caption{(a)}
			\addtocounter{figure}{-2}
			\endminipage\hfill
			\minipage{0.42\textwidth}
			\includegraphics[width=\linewidth]{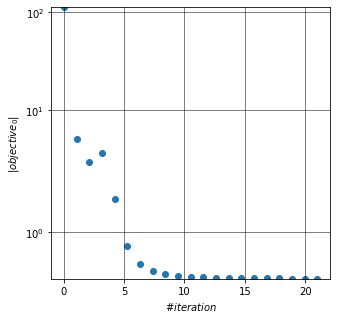}
			\captionsetup{labelformat=empty}
			\caption{(b)}
			\addtocounter{figure}{-1}
			\endminipage\hfill
			\caption{ The effect of fine-tuning on the convergence rate of the method. solving Example. \ref{test:eu_put} with a) Fine-Tuning and b) Without Fine-Tuning }
			\label{fig:fine-tuning-call_1}
		\end{figure}

		Figure. \ref{fig:optimizers} shows the superiority if Adam optimizer, as it starts to converge and moves towards the answer in earlier epochs for the first time-step. The average calculation time for SGD and RMSprop are respectively $0.038s$ and $0.045s$ per epoch.
		Figures. \ref{fig:fine-tuning-call_1}.a and \ref{fig:fine-tuning-call_1}.b illustrate the influence of fine-tuning on the objective convergence.

	\end{example}

	\section{Conclusion}\label{sec5}
	This study investigates neural networks with the famous Adam optimizer for solving financial Black-Scholes equations. Converting the PDE into a series of time-dependent ODEs using the Backward-Euler finite difference method and then solving each of these equations using the proposed model confirm the satisfactory result and fast calculation of the method. The speed of the method is caused by the parallel computations in the neural network for each independent neuron, the straight-forward calculations of sigmoid activation functions that do not add to the complexity of the model, and also the small number of training points and hidden neurons for achieving very promising accuracy. The neural network outperforms other methodologies regarding the consistency and accuracy of the model in confrontation with machines or calculation mistakes because of its fault tolerance. Fine-Tuning plays a significant role in this method by reducing the building, validation, and calculation time. It also helps the method converge faster by finding the appropriate direction for gradients as depicted in three examples in Section. \ref{sec4}. Domain Mapping, which has not been used in ANNs before to the best of found knowledge, is employed to make calculations possible on bigger or infinite problem domains. As a result of combining these approaches into one single ANN, reliable, fast, and accurate results were calculated. The methodology is applicable to other types of options priced by either ordinary or fractional models as well as other partial differential equations in any other field of study can be solved using this network.

	\bibliography{mybibfile}
	
\end{document}